\documentclass[10pt,twocolumn,letterpaper]{article}

\usepackage{cvpr}              % To produce the CAMERA-READY version
\usepackage{graphicx}
\usepackage{amsmath}
\usepackage{amssymb}
\usepackage{booktabs}
\usepackage{mathrsfs}  

\usepackage[linesnumbered,ruled,vlined]{algorithm2e}

\usepackage{times}
\usepackage[pagebackref=true,breaklinks=true,letterpaper=true,colorlinks,bookmarks=false]{hyperref}

\usepackage[utf8]{inputenc} % allow utf-8 input
\usepackage[T1]{fontenc}    % use 8-bit T1 fonts
\usepackage{url}            % simple URL typesetting
\usepackage{amsfonts}       % blackboard math symbols
\usepackage{nicefrac}       % compact symbols for 1/2, etc.
\usepackage{microtype}      % microtypography
\usepackage{subcaption}
\usepackage{xcolor}
\usepackage{bm}
\usepackage{mathtools}
\usepackage{adjustbox}
\usepackage{multirow}
\usepackage{color, colortbl}
\definecolor{mygray}{gray}{0.9}

\usepackage{helvet} 
\usepackage{courier} 
\usepackage[hyphens]{url} 
\usepackage{caption}
\DeclareCaptionStyle{ruled}{labelfont=normalfont,labelsep=colon,strut=off} % D

\usepackage{amstext}
\usepackage{enumerate}

\usepackage{newfloat}
\usepackage{listings}

\usepackage[capitalize]{cleveref}
\crefname{section}{Sec.}{Secs.}
\Crefname{section}{Section}{Sections}
\Crefname{table}{Table}{Tables}
\crefname{table}{Tab.}{Tabs.}

\def\cvprPaperID{003} % *** Enter the CVPR Paper ID here
\def\confName{CVPR}
\def\confYear{2023}

\begin{document}

%%%%%%%%% TITLE - PLEASE UPDATE
% \title{Keyword-oriented Transformer Model Improves Explainability in Ophthalmology}
% Interpretability or Interpretable
% Ophthalmologist's text annotation help model to learn Interpretability.
% \title{Keyword-driven Interpretability to Automatic Medical Report Generation}
% \title{Keyword-driven Interpretability to Automatic Medical Report Generation Models}
% \title{Keyword-driven Interpretability to Automatic Medical Report Generation Models}
% \title{ATR-ViT: Advanced Token Reduction based Vision Transformer}
% \title{Causality and Text-based Query Improve Explainability of Video Summarization}
% \title{Improving Explainability of Video Summarization by Causal Graphical Model}
% \title{Causal Explainer: Towards Explainable Automatic Video Summarization}
\title{Causalainer: Causal Explainer for Automatic Video Summarization}
% \title{$\mathcal{K}$NN-driven Token Adaption for Efficient Vision Transformer}
\author{Jia-Hong Huang$^\dagger$\\
University of Amsterdam\\
% Institution1 address\\
{\tt\small j.huang@uva.nl}
\and
Chao-Han Huck Yang\\
Georgia Institute of Technology\\
% First line of institution2 address\\
{\tt\small huckiyang@gatech.edu}
\and
Pin-Yu Chen\\
IBM Research AI\\
% First line of institution2 address\\
{\tt\small pin-yu.chen@ibm.com}
\and
Min-Hung Chen$^\ddagger$\\
NVIDIA\\
% First line of institution2 address\\
% {\tt\small vitec6@gmail.com}
{\tt\small minhungc@nvidia.com}
\and
Marcel Worring\\
University of Amsterdam\\
% First line of institution2 address\\
{\tt\small m.worring@uva.nl}
}

% \author{First Author\\
% Institution1\\
% Institution1 address\\
% {\tt\small firstauthor@i1.org}
% For a paper whose authors are all at the same institution,
% omit the following lines up until the closing ``}''.
% Additional authors and addresses can be added with ``\and'',
% just like the second author.
% To save space, use either the email address or home page, not both
% \and
% Second Author\\
% Institution2\\
% First line of institution2 address\\
% {\tt\small secondauthor@i2.org}
% }
\maketitle

% \twocolumn[{%
% \renewcommand\twocolumn[1][]{#1}%
% \maketitle
% \begin{center}
%     \centering
%     \captionsetup{type=figure}
%     \includegraphics[width=1.0\linewidth,height=6.5cm]{discover.pdf}
%     \vspace{-0.50cm}
%     \captionof{figure}{Visualization of human-annotated and machine-predicted frame-level scores for creating a video summary. Comparing the human-annotated video summary score pattern to the one generated by existing state-of-the-art video summarization methods, e.g., \cite{vasudevan2017query,huang2020query,huang2021gpt2mvs}, we observe these methods are capable of learning visual consecutiveness and diversity which are some key factors considered by humans for creating a good video summary. These methods mainly focus on capturing the visual cues to achieve such a purpose. Red bars denote discarded frames and grey bars indicate selected frames used to form a video summary. The video has $199$ frames and the numbers, except for $199$, denote the indices of frames.}
% \label{fig:figure21}
% \end{center}%
% }]

\begin{abstract}
The goal of video summarization is to automatically shorten videos such that it conveys the overall story without losing relevant information. In many application scenarios, improper video summarization can have a large impact. For example in forensics, the quality of the generated video summary will affect an investigator's judgment while in journalism it might yield undesired bias. Because of this, modeling explainability is a key concern. One of the best ways to address the explainability challenge is to uncover the causal relations that steer the process and lead to the result. Current machine learning-based video summarization algorithms learn optimal parameters but do not uncover causal relationships. Hence, they suffer from a relative lack of explainability. In this work, a Causal Explainer, dubbed Causalainer, is proposed to address this issue. Multiple meaningful random variables and their joint distributions are introduced to characterize the behaviors of key components in the problem of video summarization. In addition, helper distributions are introduced to enhance the effectiveness of model training. In visual-textual input scenarios, the extra input can decrease the model performance. A causal semantics extractor is designed to tackle this issue by effectively distilling the mutual information from the visual and textual inputs. Experimental results on commonly used benchmarks demonstrate that the proposed method achieves state-of-the-art performance while being more explainable. 
\end{abstract}

\let\thefootnote\relax\footnotetext{\scriptsize{{$^\dagger$Work done during an internship at Microsoft Research in Cambridge, UK and Amsterdam, NL. $^\ddagger$ex-Microsoft.}}}
% \let\thefootnote\relax\footnotetext{\scriptsize{{$^\dagger$Work done during an internship at Microsoft Research, Cambridge, UK.}}}

%%%%%%%%% BODY TEXT
\section{Introduction}
\label{section:section1}

Video summarization is the process of automatically generating a concise video clip that conveys the primary message or story in the original video. Various automatic video summarization algorithms have been proposed in recent years to tackle this task using different supervision schemes. These include fully-supervised methods that utilize visual input alone \cite{gong2014diverse,gygli2014creating,zhang2016video,zhao2017hierarchical,zhao2018hsa,zhang2019dtr,ji2019video,ji2020deep} or multi-modal input \cite{li2017extracting,vasudevan2017query,sanabria2019deep,song2016category,zhou2018video,lei2018action,otani2016video,yuan2017video,wei2018video,huang2020query,huang2021gpt2mvs}, as well as weakly-supervised methods \cite{panda2017weakly,ho2018summarizing,cai2018weakly,chen2019weakly,jiang2019comprehensive,yan2020self}.

\begin{figure}[t!]
\begin{center}
\includegraphics[width=0.9\linewidth]{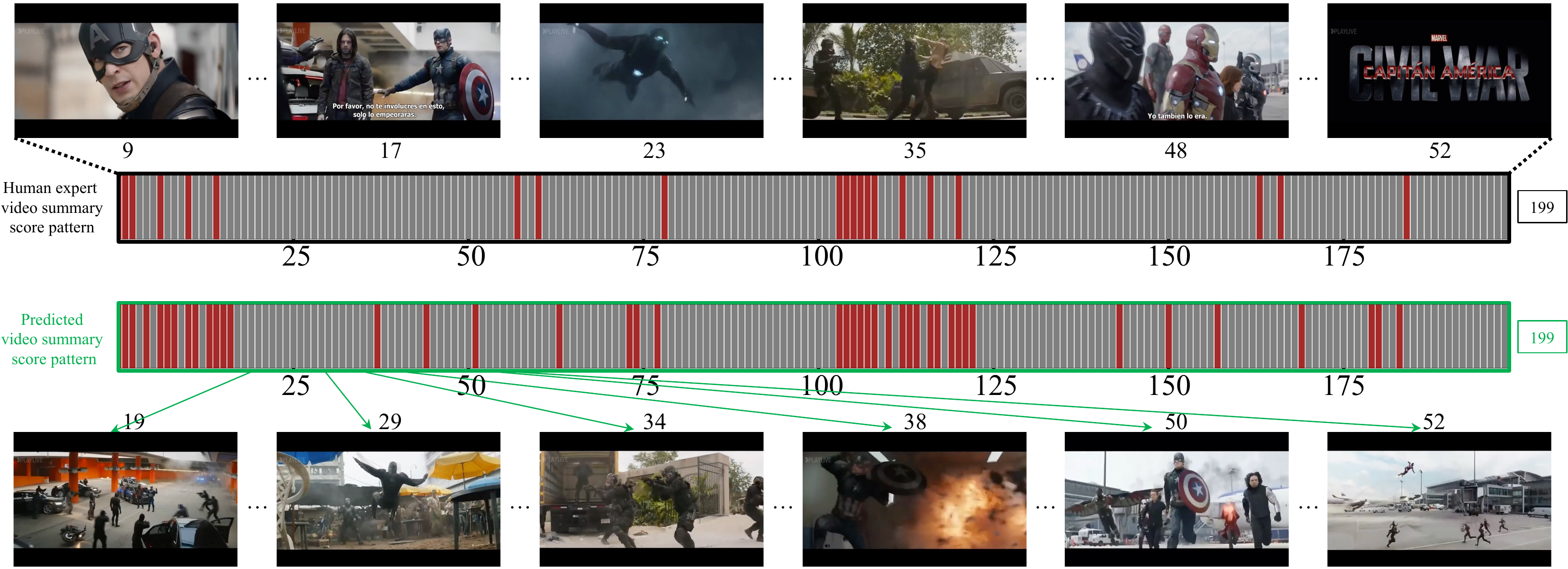}
\end{center}
\vspace{-0.60cm}
   \caption{Visualization of human-annotated and machine-predicted frame-level scores for creating a video summary. Comparing the human-annotated video summary score pattern to the one generated by existing state-of-the-art video summarization methods, e.g., \cite{vasudevan2017query,huang2020query,huang2021gpt2mvs}, we observe these methods are capable of learning visual consecutiveness and diversity which are some key factors considered by humans for creating a good video summary. These methods mainly focus on capturing the visual cues to achieve such a purpose. Red bars denote discarded frames and grey bars indicate selected frames used to form a summary. The video has $199$ frames and the numbers, except for $199$, denote the indices of frames.
   }
\vspace{-0.6cm}
\label{fig:figure21}
\end{figure}

According to \cite{gygli2015video,song2015tvsum,vasudevan2017query,gong2014diverse,gygli2014creating,huang2020query}, when human experts perform the task of video summary generation, they will not only consider concrete/visual factors, e.g., visual consecutiveness and visual diversity, but also abstract/non-visual factors, such as interestingness, representativeness, and storyline smoothness. Hence, a human-generated video summary is based on many confounding factors. These factors/causes result in the video summary. Existing works do not, or in a very limited way, consider abstract factors and mainly focus on proposing various methods to exploit concrete visual cues to perform video summarization. See the illustration in Figure \ref{fig:figure21}. This leads to limited modeling explainability of automatic video summarization \cite{huang2020query,jiang2022joint}. 

Machine learning (ML) models can be made more explainable through causation modeling based on Bayesian probability \cite{pearl2001bayesianism,pearl2018theoretical,zhang2020causal,khemakhem2021causal,zhang2018advances}. In this work, we propose a novel method for improving the inherent explainability of video summarization models called Causalainer, which is based on causation modeling. Our approach aims to address the challenge of model explainability in video summarization by leveraging the insights gained from Bayesian probability and causation modeling. See Figure \ref{fig:figure0} for the method flowchart of the proposed Causalainer. To model the problem of video summarization and increase the explainability, four meaningful random variables are introduced to characterize the behaviors of the data intervention, the model's prediction, observed potential confounders, and unobserved confounders, respectively. Note that data intervention is a way to help a model learn the causal relations that lead to the result \cite{louizos2017causal,greenland1983correcting,selen1986adjusting,cai2012identifying,greenland2011bias,shalit2017estimating}. A prior joint distribution and its posterior approximation can be built on top of those four random variables. The proposed method is trained based on minimizing the distance between the prior distribution and the posterior approximation. We identify that predicting the behaviors of the data intervention and model's outcome can be challenging in practice due to various factors, e.g., video noise, lens or motion blur. We address this issue by introducing helper distributions for them. The helper distributions form a new loss term to guide the model learning. Furthermore, when multi-modal inputs are available, we identify that the extra input sometimes can harm the model performance most likely due to the interactions between different modalities being ineffective. We address this challenge by introducing a causal semantics extractor to effectively distill the mutual information between multi-modal inputs.

These novel design choices have been instrumental in improving the explainability and performance of video summarization models. The extensive experimentation on commonly used video summarization datasets verifies that the proposed method outperforms existing state-of-the-art while also providing greater explainability. By leveraging causal learning techniques, our approach represents a promising attempt to reinforce the causal inference ability and explainability of an ML-based video summarization model.

\section{Methodology}
\label{methodology:method}
\begin{figure}[t!]
\begin{center}
\includegraphics[width=0.9\linewidth]{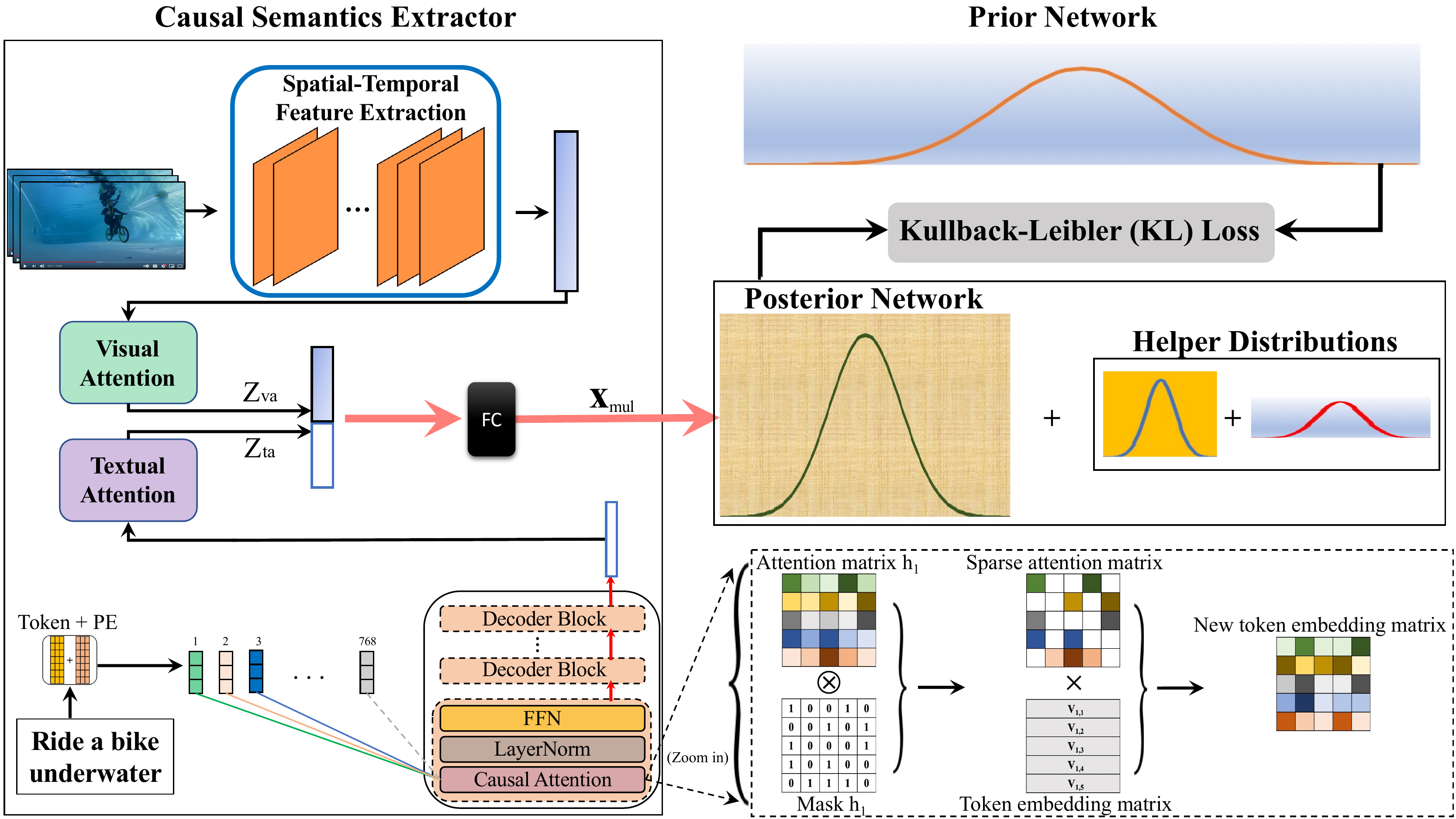}
\end{center}
\vspace{-0.60cm}
   \caption{Flowchart of the proposed Causal Explainer (Causalainer) method for video summarization. The proposed method is mainly composed of a prior network, a posterior network, helper distributions, and a causal semantics extractor. $\otimes$ denotes element-wise multiplication and $\times$ indicates matrix multiplication. ``Token + PE'' denotes the operations of token embedding and positional encoding.
   }
\vspace{-0.5cm}
\label{fig:figure0}
\end{figure}

% \noindent\textbf{3.1 Overview}
We now present the details of the proposed Causal Explainer method for automatic video summarization, dubbed Causalainer. First, the assumptions of causal modeling are described in detail. Secondly, we introduce four random variables $\textbf{y}$, $\textbf{t}$, $\textbf{X}$, and $\textbf{Z}$ to characterize the behaviors of the model's prediction, the data intervention, observed potential confounders, and unobserved confounders, respectively. Finally, the derivation of our training objective with helper distributions and the proposed causal semantics extractor are presented. Causalainer consists of prior and posterior probabilistic networks. See Figure \ref{fig:figure0} for an overview.

% The proposed Causalainer approach is mainly composed of two probabilistic networks, namely the prior and posterior networks. An overview of the proposed method is presented in Figure \ref{fig:figure0}. 

% In this section, the proposed approach and assumptions of causal learning for video summarization are described in detail. The proposed method is based on causal learning with an interactive causal attention mechanism, contextualized query representations, convolutional 3D (C3D) features, and video summary generation. An overview of the proposed method is presented in Figure \ref{fig:figure0}.
% % and it is a practical implementation of a causal graphical model in video summarization, referring to \textbf{Supplementary Section A}. 
% The proposed attentive causal learning helps increase a video summarization model's causal inference ability. The goal of the interactive causal attention is to capture the interactive information in the components of the proposed method. The contextualized query representations are based on the architecture of GPT-2. C3D captures the spatial and temporal information of an input video. The proposed method simultaneously generates outcome score labels $\textbf{y}$ and a reconstruction of $\textbf{X}$. Thereafter, the video summary generator creates summaries based on $\textbf{y}$. 

% \subsection{Assumptions}
\noindent\textbf{3.1 Assumptions}

In general, causal learning for real-world observational studies is complicated \cite{louizos2017causal,abbasnejad2020counterfactual,yang2021causal,agarwal2020towards,huang2019novel,huang2017vqabq,huang2017robustness,huang2017robustnessMS,huang2023improving,huang2019assessing,huang2021deepopht,huang2022non,huang2021contextualized,huang2021deep,huang2021longer,hu2019silco,huck2018auto,liu2018synthesizing,yang2018novel,yang2023treatment,di2021dawn,wu2023expert,huang2022causal}. With the established efforts \cite{pearl2018theoretical,louizos2017causal,zhang2018advances} on causal learning under noisy interventions, two assumptions are imposed when modeling the problem of video summarization. 
% based on a causal graphical model, referring to \textbf{Supplementary Section A}. 
First, the information of having visual/textual intervention $\textbf{t}$ or not is binary. Second, the observations $(\textbf{X}, \textbf{t}, \textbf{y})$ from a deep neural network (DNN) are sufficient to approximately recover the joint distribution $p(\textbf{Z}, \textbf{X}, \textbf{t}, \textbf{y})$ of the unobserved/latent confounding variable $\textbf{Z}$, the observed confounding variable $\textbf{X}$, the intervention $\textbf{t}$, and the outcome $\textbf{y}$. The proposed Causalainer method is built on top of multiple probability distributions as described in the following subsections.

% \subsection{Causal Explainer for Video Summarization}
\noindent\textbf{3.2 Causal Explainer for Video Summarization}

% According to \cite{kingma2013auto,rezende2014stochastic,kingma1606improving,rezende2015variational,tran2015variational,louizos2017causal}, it has been proved that variational autoencoders (VAE) are capable of learning a latent variable causal graphical model. Based on \cite{louizos2017causal,kingma2013auto}, VAE aim to learn a variational latent representation $\textbf{Z}$ from data $\textbf{X}$ for reconstruction. Moreover, the authors of \cite{sonderby2016ladder,chen2016variational,tomczak2017vae} have recently expanded the range and type of distributions that can be captured by VAE. Hence, to effectively infer the complex non-linear relationships between $\textbf{X}$ and $(\textbf{Z}, \textbf{t}, \textbf{y})$ and to approximately recover $p(\textbf{Z}, \textbf{X}, \textbf{t}, \textbf{y})$, the proposed attentive causal learning for video summarization is built on top of VAE. The proposed causal learning is composed of a decoder and an encoder. According to \cite{louizos2017causal}, the decoder can be considered as a model network. Based on \cite{kingma2013auto,rezende2014stochastic}, the encoder can be considered as an inference network. 
In the proposed Causalainer, $\textbf{x}_i$ denotes an input video and an optional text-based query indexed by $i$, $\textbf{z}_i$ indicates the latent confounder, $t_i \in \{0,1\}$ denotes the intervention assignment, and $y_i$ indicates the outcome.

\noindent\textbf{Prior Probability Distributions.}
% The proposed Causalainer approach is mainly composed of two probabilistic networks, dubbed prior network and posterior network.
The prior network is conditioning on the latent variable $\textbf{z}_i$ and mainly consists of the following components: \textbf{(i)} The latent confounder distribution: 
$p(\textbf{z}_i) = \prod_{z\in \textbf{z}_i} \mathcal{N}(z | \mu=0, \sigma^2=1),$ 
% \begin{align}
%     p(\textbf{z}_i) = \prod_{z\in \textbf{z}_i} \mathcal{N}(z | \mu=0, \sigma^2=1),
% \label{eq:eq1}
% \end{align}
where $\mathcal{N}(z | \mu, \sigma^2)$ denotes a Gaussian distribution with a random variable $z$, $z$ is an element of $\textbf{z}_i$, and the mean $\mu$ and variance $\sigma^2$ follow the settings in \cite{kingma2013auto}, i.e., $\mu=0$ $\sigma^2=1$.
\textbf{(ii)} The conditional data distribution:
$p(\textbf{x}_i | \textbf{z}_i) = \prod_{x \in \ \textbf{x}_i} p(x | \textbf{z}_i),$
% \begin{align}
%     p(\textbf{x}_i | \textbf{z}_i) = \prod_{x \in \ \textbf{x}_i} p(x | \textbf{z}_i),
% \label{eq:eq2}
% \end{align}
where $p(x |\textbf{z}_i)$ is an appropriate probability distribution with a random variable $x$, the distribution is conditioning on $\textbf{z}_i$, and $x$ is an element of $\textbf{x}_i$.
\textbf{(iii)} The conditional intervention distribution:
$p(t_i | \textbf{z}_i) = \textup{Bernoulli} (\sigma(f_{\theta_1}(\textbf{z}_i))),$
% \begin{align}
%     p(t_i | \textbf{z}_i) = \textup{Bernoulli} (\sigma(f_{\theta_1}(\textbf{z}_i))),
% \label{eq:eq3}
% \end{align}
where $\sigma(\cdot)$ is a logistic function, $\textup{Bernoulli}  (\cdot)$ indicates a Bernoulli distribution for a discrete outcome, and $f_{\theta_1}(\cdot)$ denotes a neural network parameterized by the parameter $\theta_1$.
\textbf{(iv)} The conditional outcome distribution:
$p(y_i | \textbf{z}_i, t_i) = \sigma(t_{i} f_{\theta_2}(\textbf{z}_i) + (1 - t_{i}) f_{\theta_3}(\textbf{z}_i)),$
% \begin{align}
%     p(y_i | \textbf{z}_i, t_i) = \sigma(t_{i} f_{\theta_2}(\textbf{z}_i) + (1 - t_{i}) f_{\theta_3}(\textbf{z}_i)),
% \label{eq:eq4}
% \end{align}
where $f_{\theta_2}(\cdot)$ and $f_{\theta_3}(\cdot)$ are neural networks parameterized by the parameters $\theta_2$ and $\theta_3$, respectively.
In this work, $y_i$ is tailored for a categorical classification problem, i.e., frame-based importance score classification in video summarization.
% , but the formulation can be naturally extended to different tasks. For example, for regression tasks, one can simply remove the logistic function $\sigma(\cdot)$ in $p(y_i | \textbf{z}_i, t_i)$. 

\noindent\textbf{Posterior Probability Distribution.}
Since a priori knowledge on the latent confounder does not exist, we have to marginalize over it in order to learn the model parameters, $\theta_1$, $\theta_2$, and $\theta_3$ in \textbf{(iii)} and \textbf{(iv)}. The non-linear neural network functions make inference intractable. 
Hence, variational inference \cite{kingma2013auto} along with the posterior network is employed.
% , referring to Figure \ref{fig:figure1}-(a) and Figure \ref{fig:figure1}-(b). 
These neural networks output the parameters of a fixed form posterior approximation over the latent variable $\textbf{z}$, given the observed variables. 
Similar to \cite{louizos2017causal,rezende2014stochastic}, in this work, the proposed posterior network is conditioning on observations. Also, the true posterior over $\textbf{Z}$ depends on $\textbf{X}$, $\textbf{t}$ and $\textbf{y}$.
% , referring to \textbf{Supplementary Section A} for more details. 
Hence, the posterior approximation defined below is employed to build the posterior network.
$q(\textbf{z}_i | \textbf{x}_i, y_i, t_i) = \prod_{z\in \textbf{z}_i} \mathcal{N}(z | \bm{\mu}_i, \bm{\sigma^2}_i),$ 
% \begin{align}
%     q(\textbf{z}_i | \textbf{x}_i, y_i, t_i) = \prod_{z\in \textbf{z}_i} \mathcal{N}(z | \bm{\mu}_i, \bm{\sigma^2}_i),
% \label{eq:eq5}
% \end{align}
where $\bm{\mu}_i = t_i \bm{\mu}_{t=1, i} + (1 - t_i) \bm{\mu}_{t=0,i}$, $\bm{\sigma^2}_i = t_i \bm{\sigma^2}_{t=1, i} + (1 - t_i) \bm{\sigma^2}_{t=0, i}$, $\bm{\mu}_{t=0, i} = g_{\phi_1} \circ g_{\phi_0}(\textbf{x}_i, y_i)$, $\bm{\sigma^2}_{t=0, i} = \sigma(g_{\phi_2} \circ g_{\phi_0}(\textbf{x}_i, y_i))$, $\bm{\mu}_{t=1, i} = g_{\phi_3} \circ g_{\phi_0}(\textbf{x}_i, y_i)$, $\bm{\sigma^2}_{t=1, i} = \sigma(g_{\phi_4} \circ g_{\phi_0}(\textbf{x}_i, y_i))$, $g_{\phi_k}(\cdot)$ denotes a neural network with variational parameters $\phi_k$ for $k=0, 1, 2, 3, 4$, and $g_{\phi_0}(\textbf{x}_i, y_i)$ is a shared representation. Note that a feature map is multiplied with the approximated posterior $q(y_i | \textbf{x}_i, t_i)$ without logistic function $\sigma(\cdot)$ to get $g_{\phi_0}(\textbf{x}_i, y_i)$.
% , referring to Figure \ref{fig:figure1}-(a).

% \subsection{Training Objective with Helper Distributions}
\noindent\textbf{3.3 Training Objective with Helper Distributions}

In practice, various factors, e.g., video noise, motion blur, or lens blur, make the prediction of the behaviors of the data intervention and the model’s outcome challenging. Therefore, two helper distributions are introduced to alleviate this issue.
% The true posterior over $\textbf{Z}$ depends not only on $\textbf{X}$, but also on $\textbf{t}$ and $\textbf{y}$.
We have to know the intervention assignment $\textbf{t}$ along with its outcome $\textbf{y}$ before inferring the distribution over $\textbf{Z}$. Hence, the helper distribution 
% defined below is introduced for the intervention assignment $t_{i}$.
$q(t_i | \textbf{x}_i) = \textup{Bernoulli}(\sigma(g_{\phi_5}(\textbf{x}_i)))$ is introduced for the intervention assignment $t_{i}$, and  
% \begin{align}
%     q(t_i | \textbf{x}_i) = \textup{Bernoulli}(\sigma(g_{\phi_5}(\textbf{x}_i))).
% \end{align}
the other helper distribution 
% defined below is introduced for the outcome $y_{i}$.
$q(y_i | \textbf{x}_i, t_i) = \sigma(t_{i} g_{\phi_6}(\textbf{x}_i) + (1 - t_{i}) g_{\phi_7}(\textbf{x}_i))$ is introduced for the outcome $y_{i}$,
% \begin{align}
%     q(y_i | \textbf{x}_i, t_i) = \sigma(t_{i} g_{\phi_6}(\textbf{x}_i) + (1 - t_{i}) g_{\phi_7}(\textbf{x}_i)),
% \end{align}
where $g_{\phi_k}(\cdot)$ indicates a neural network with variational parameters $\phi_k$ for $k=5, 6, 7$. 
% \noindent Note that traditional VAE simply passes the feature map directly to the latent space, i.e., the upper path of the proposed PEM in Figure \ref{fig:figure1}-(a). To increase the interaction between the feature map and the approximated posterior distribution, in this work the feature map is also sent to the middle and the lower paths of the PEM in Figure \ref{fig:figure1}-(a)) for the posterior estimations of the intervention $t_i$ and the outcome $y_i$. \\
The introduced helper distributions benefit the prediction of $t_i$ and $y_i$ for new samples. To estimate the variational parameters of the distributions $q(t_i | \textbf{x}_i)$ and $q(y_i | \textbf{x}_i, t_i)$, a helper objective function $\mathcal{L}_{\textup{helper}} = \sum^{N}_{i=1}[ \log q(t_i=t_i^{*} | \textbf{x}_i^{*}) + \log q(y_i=y_i^{*} | \textbf{x}_i^{*}, t_i^{*})]$ is introduced to the final training objective over $N$ data samples, where $\textbf{x}_i^*$, $t_i^*$ and $y_i^*$ are the observed values in the training set.
% \textcolor{red}{Now, it is time to form a single objective for the encoder, in Figure \ref{fig:figure2}-(a), and the decoder, in Figure \ref{fig:figure2}-(b), to learn meaningful causal representations in the latent space and generate video summaries.}
The overall training objective $\mathcal{L}_{\textup{causal}}$ for the proposed method is defined below.
% The variational lower bound, \cite{kingma2013auto,rezende2014stochastic,louizos2017causal}, of our causal graphical model, CAVSM, to be optimized is given by:
$\mathcal{L}_{\textup{causal}} = \mathcal{L}_{\textup{helper}} + \nonumber
\sum^{N}_{i=1}\mathbb{E}_{q(\textbf{z}_{i}|\textbf{x}_{i}, t_i, y_i)}[\log p(\textbf{x}_i, t_i | \textbf{z}_i) + \nonumber
\log p(y_i|t_i, \textbf{z}_i) + \log p(\textbf{z}_i) - \log q(\textbf{z}_i|\textbf{x}_i, t_i, y_i)].$
% \begin{equation}
%     \mathcal{L}_{\textup{causal}} = \mathcal{L}_{\textup{helper}} ~ + \nonumber
    % \mathcal{L}_{causal} = \sum^{N}_{i=1}[( \log q(t_i=t_i^{*} | \textbf{x}_i^{*}) + \log q(y_i=y_i^{*} | \textbf{x}_i^{*}, t_i^{*}))] \nonumber
% \end{equation}
% \vspace{-0.4cm}
% \begin{equation}
%     \sum^{N}_{i=1}\mathbb{E}_{q(\textbf{z}_{i}|\textbf{x}_{i}, t_i, y_i)}[\log p(\textbf{x}_i, t_i | \textbf{z}_i) ~ + \nonumber
    % \log p(y_i|t_i, \textbf{z}_i) + \log p(\textbf{z}_i) - \log q(\textbf{z}_i|\textbf{x}_i, t_i, y_i)].
    % \label{equ_15}
% \end{equation}
% \begin{equation}
%     % \sum^{N}_{i=1}\mathbb{E}_{q(\textbf{z}_{i}|\textbf{x}_{i}, t_i, y_i)}[\log p(\textbf{x}_i, t_i | \textbf{z}_i) + 
%     \log p(y_i|t_i, \textbf{z}_i) + \log p(\textbf{z}_i) - \log q(\textbf{z}_i|\textbf{x}_i, t_i, y_i)].
%     \label{equ_15}
% \end{equation}
% Please refer to \cite{odaibo2019tutorial} and \textbf{Supplementary Section D} for the detailed derivation of the evident lower bound (ELBO) $ELBO =  \mathcal{L}_{causal} - \mathcal{L}_{helper}$ \cite{kingma2013auto,rezende2014stochastic,louizos2017causal} of the VAE-based causal graphical model in video summarization. 

% \textcolor{red}{Please refer to \textbf{Supplementary Section \ref{supplementary:section-4}} for the detailed derivation of the evident lower bound (ELBO) $ELBO =  \mathcal{L}_{causal} - \mathcal{L}_{helper}$ \cite{kingma2013auto,rezende2014stochastic,louizos2017causal} of the causal graphical model in video summarization.} 

% \noindent\textbf{3.4 Training Objective}

% \subsection{Causal Semantics Extractor}
\noindent\textbf{3.4 Causal Semantics Extractor}

Existing commonly used video summarization datasets, e.g., TVSum \cite{song2015tvsum} and QueryVS \cite{huang2020query}, provide visual and textual inputs. Since the textual input cannot always help the model performance because of the ineffective extraction of mutual information from the visual and textual inputs, a causal semantics extractor is introduced to alleviate this issue. The proposed extractor is built on top of transformer blocks \cite{vaswani2017attention}. Vanilla transformers exploit all of the tokens in each layer for attention computation. However, the design philosophy of the proposed causal semantics extractor, dubbed causal attention, is effectively using fewer but relatively informative tokens to compute attention maps, instead of using the total number of tokens. According to \cite{vaswani2017attention}, the computation of the vanilla attention matrix $\mathscr{A} \in \mathbb{R}^{n\times n}$ is based on the dot-product. It is defined as 
$\mathscr{A} = \textup{softmax}\left ( \frac{\mathbf{Q}\mathbf{K}^\top}{\sqrt{d}} \right );
    \mathbf{Q} = \mathbf{T}\mathbf{W}_{q},
    \mathbf{K} = \mathbf{T}\mathbf{W}_{k},$
% in Equation (\ref{eq:eq11}).
% \begin{align}
%     % Q &= W_q \phi(I)+b_q \\
%     \mathscr{A} &= \textup{softmax}\left ( \frac{\mathbf{Q}\mathbf{K}^\top}{\sqrt{d}} \right );
%     \mathbf{Q} = \mathbf{T}\mathbf{W}_{q},
%     \mathbf{K} = \mathbf{T}\mathbf{W}_{k},
%     \label{eq:eq11}
% \end{align}
where the query matrix $\mathbf{Q} \in \mathbb{R}^{n\times d}$ and key matrix $\mathbf{K} \in \mathbb{R}^{n\times d}$ are generated by the linear projection of the input token matrix $\mathbf{T} \in \mathbb{R}^{n\times d_{m}}$ based on the learnable weights matrices $\mathbf{W}_{q} \in \mathbb{R}^{d_{m}\times d}$ and $\mathbf{W}_{k} \in \mathbb{R}^{d_{m}\times d}$. $n$ indicates the total number of input tokens. $d$ represents the embedding dimension and $d_{m}$ denotes the dimension of an input token. 
The new value matrix $\mathbf{V}_{\textup{new}} \in \mathbb{R}^{n\times d}$ can be obtained via 
% Equation (\ref{eq:eq22}).
$\mathbf{V}_{\textup{new}} = \mathscr{A}\mathbf{V};
   \mathbf{V} = \mathbf{T}\mathbf{W}_{v},$
% \begin{align}
%     \mathbf{V}_{\textup{new}} = \mathscr{A}\mathbf{V};
%     \mathbf{V} = \mathbf{T}\mathbf{W}_{v},
%     \label{eq:eq22}
% \end{align}
where the value matrix $\mathbf{V} \in \mathbb{R}^{n\times d}$ and $\mathbf{W}_{v} \in \mathbb{R}^{d_{m}\times d}$.

In \cite{vaswani2017attention}, the vanilla attention matrix is based on the calculation of all the query-key pairs. However, in the proposed Causal Semantics Extractor, only the top $\kappa$ most similar keys and values for each query are used to compute the causal attention matrix. Similar to \cite{vaswani2017attention}, all the queries and keys are calculated by the dot-product. Then, the row-wise top $\kappa$ elements are used for the \textup{softmax} calculation. In the proposed Causal Semantics Extractor, the value matrix $\mathbf{V}_{\kappa} \in \mathbb{R}^{n\times d}$ is defined as 
$  \mathbf{V}_{\kappa} =\textup{softmax}\left (\tau _{\kappa} (\mathbf{\mathscr{A}}) \right ) \mathbf{V}_{\textup{new}}  \nonumber 
    = \textup{softmax}\left (\tau _{\kappa} \left ( \frac{\mathbf{Q}\mathbf{K}^\top}{\sqrt{d}} \right )\right )\mathbf{V}_{\textup{new}},$
% in Equation (\ref{eq:eq33}).
% \begin{align}
%     \mathbf{V}_{\kappa} =\textup{softmax}\left (\tau _{\kappa} (\mathbf{\mathscr{A}}) \right ) \mathbf{V}_{\textup{new}}  \nonumber \\
%     = \textup{softmax}\left (\tau _{\kappa} \left ( \frac{\mathbf{Q}\mathbf{K}^\top}{\sqrt{d}} \right )\right )\mathbf{V}_{\textup{new}},
%     \label{eq:eq33}
% \end{align}
where $\tau _{\kappa}(\cdot)$ denotes an operator for the row-wise top $\kappa$ elements selection. $\tau _{\kappa}(\cdot)$ is defined as
$[\tau _{\kappa}(\mathbf{\mathscr{A}})]_{ij}=\begin{cases}
\mathscr{A}_{ij} &, \mathscr{A}_{ij}\in \text{top $\kappa$ factors at row~$i$} \\ 
-\infty  &, \text{ otherwise}.
\end{cases}$
% \begin{align}
% [\tau _{\kappa}(\mathbf{\mathscr{A}})]_{ij}=\begin{cases}
% \mathscr{A}_{ij} &, \mathscr{A}_{ij}\in \text{top $\kappa$ factors at row~$i$} \\ 
% -\infty  &, \text{ otherwise}.
% \end{cases}
% \label{eq:eq44}
% \end{align}
Then, $\mathbf{V}_{\kappa}$ can be further used to generate $\mathbf{X}_{\textup{mul}}$, i.e.,  an output of the proposed Causal Semantics Extractor. The procedure for calculating $\mathbf{X}_{\textup{mul}}$ is defined below.
$Z_{\textup{ta}} = \textup{\textup{TextAtten}}(\textup{FFN}(\textup{LayerNorm}(\mathbf{V}_{\kappa})),$
% \begin{equation}
%     Z_{\textup{ta}} = \textup{\textup{TextAtten}}(\textup{FFN}(\textup{LayerNorm}(\mathbf{V}_{\kappa})),
%     \label{eq:zta}
% \end{equation}
where $\textup{LayerNorm}(\cdot)$ denotes a layer normalization, $\textup{FFN}(\cdot)$ indicates a feed forward network, and $\textup{TextAtten}(\cdot)$ denotes an element-wise multiplication-based textual attention mechanism.
$Z_{\textup{va}} = \textup{VisualAtten}(\textup{C3D}(\mathbf{I})),$
% \begin{equation}
%     Z_{\textup{va}} = \textup{VisualAtten}(\textup{C3D}(\mathbf{I})),
%     \label{eq:zva}
% \end{equation}
where $I$ denotes an input video, $\textup{C3D}(\cdot)$ indicates an operation of the spatial-temporal feature extraction, e.g., 3D version of ResNet-34 \cite{he2016deep,hara2018can}, for the input video, and $\textup{VisualAtten}(\cdot)$ indicates a visual attention mechanism based on the element-wise multiplication.
$\mathbf{X}_{\textup{mul}} = \textup{FC}(Z_{\textup{ta}} \odot Z_{\textup{va}}),$
% \begin{equation}
%     \mathbf{X}_{\textup{mul}} = \textup{FC}(Z_{\textup{ta}} \odot Z_{\textup{va}}),
%     \label{eq:xmul}
% \end{equation}
where $\odot$ denotes the operation of feature concatenation and $\textup{FC}(\cdot)$ indicates a fully connected layer. Note that the Causal Semantics Extractor's output $\mathbf{X}_{\textup{mul}}$ is an input of the proposed posterior network based on the scheme of using multi-modal inputs.
% , and $\mathbf{X}_{\textup{mul}}$ indicates an input of the proposed posterior network based on the scheme of using multi-modal inputs.

Similar to the final step of video summary generation in \cite{huang2020query}, after the end-to-end training of the proposed causal video summarization model is complete, the trained model can be used for video summary generation. Finally, based on the generated score labels, a set of video frames is selected from the original input video to form a final video summary. Note that the summary budget is considered as a user-defined hyper-parameter in multi-modal video summarization \cite{huang2020query}.

\section{Experiments}
\noindent\textbf{3.1 Experimental Setup and Datasets Preparation}
\label{section:section4-1}

\noindent\textbf{Experimental Setup.}
We consider three scenarios: 1) fully-supervised training with human-defined frame-level labels, 2) fully-supervised training with multi-modal input including text-based query, and 3) weakly-supervised learning with two-second segment-level scores, which can be considered as a form of weak label \cite{cai2018weakly,chen2019weakly,apostolidis2021video}. Note that \cite{song2015tvsum} empirically finds that a two-second segment length is appropriate for capturing video local context with good visual coherence. Hence, in this work, a video segment-level score is produced per two seconds based on given frame-level scores.

% In this work, the following three scenarios are considered. First, in a fully-supervised scheme, a full set of data with human expert annotations, i.e., frame-level labels, are used to train the proposed model. Secondly, in a fully-supervised scenario with the multi-modal input, the text-based query is considered as an input. Third, the authors of \cite{song2015tvsum} empirically find that a two-second segment length is appropriate for capturing video local context with good visual coherence. Hence, a video segment-level score is produced per two seconds based on given frame-level scores. The segment-level label can be considered as a type of weak label in a weakly-supervised learning scheme \cite{cai2018weakly,chen2019weakly,apostolidis2021video}. 

\noindent\textbf{Video Summarization Datasets.}
In the experiments, three commonly used video summarization datasets, i.e., TVSum \cite{song2015tvsum}, QueryVS \cite{huang2020query}, and SumMe \cite{gygli2014creating}, are exploited to evaluate the proposed method. The TVSum dataset contains $50$ videos.
% The TVSum dataset is randomly divided into 40/5/5 videos for training/validation/testing, respectively. 
The length of the video in TVSum is ranging from $2$ to $10$ minutes. The human expert frame-level importance score label in TVSum is ranging from $1$ to $5$. The QueryVS dataset contains $190$ videos.
% The QueryVS dataset is separated into 114/38/38 videos for training/validation/testing, respectively. 
The video length in QueryVS is ranging from $2$ to $3$ minutes. The human expert frame-level importance score label in QueryVS is ranging from $0$ to $3$. Every video is retrieved based on a given text-based query. The SumMe dataset contains $25$ videos.
% The entire SumMe dataset is randomly divided into 19/3/3 videos for training/validation/testing, respectively. 
The video duration in SumMe is ranging from $1$ to $6$ minutes. In SumMe, the importance score annotated by human experts ranges from $0$ to $1$. Note that SumMe is not used for multi-modal video summarization. Hence, we do not have textual input when a model is evaluated on this dataset. Videos from these datasets are sampled at $1$ frame per second (fps). The input image size is $224$ by $224$ with RGB channels. Every channel is normalized by standard deviation $=(0.2737, 0.2631, 0.2601)$ and mean $=(0.4280, 0.4106, 0.3589)$. PyTorch and NVIDIA TITAN Xp GPU are used for the implementation and to train models for $60$ epochs with $1e-6$ learning rate. The Adam optimizer is used \cite{kingma2014adam}, with hyper-parameters set as $\epsilon=1e-8$, $\beta_{1}=0.9$, and $\beta_{2}=0.999$. 

\noindent\textbf{Causal Learning Dataset.}
\label{section:section3-5-1}
When we observe people's writing behaviors, we notice some of them happen very often, such as synonym replacement, accidentally missing some words in a sentence, and so on. Motivated by the above, we randomly pick up one of the behaviors, e.g., accidentally missing some words in a sentence, and write a textual intervention function to simulate it. Similarly, we know that when people make videos in their daily life, some visual disturbances may exist, e.g., salt and pepper noise, image masking, blurring, and so on. We also randomly pick up some of them, e.g., blur and salt and pepper noise, and make a visual intervention function to do the simulation. Based on the visual and textual simulation functions, we can make our causal video summarization dataset with visual and textual interventions. The dataset is made based on the following steps. First, 50\% of the \textit{(video, query)} data pairs are randomly selected from the original training, validation, and testing sets. Secondly, for each selected video, $0$ or $1$ intervention labels are randomly assigned to $30$\% of the video frames and the corresponding queries. 
% Figure \ref{fig:figure5} and Figure \ref{fig:figure6} are dataset examples with visual intervention and textual intervention, respectively. 
Note that in real-world scenarios, there are various disturbances beyond the previously mentioned visual and textual interventions that could be utilized in the proposed method.
% Finally, the same method mentioned in \cite{huang2020query,sharghi2018improving} is used to make each video has the same length.

\begin{table}[t!]
    \caption{Comparison with fully-supervised state-of-the-art methods. The proposed method performs the best on both datasets. Note that textual query input is not used in this experiment.
    }
\vspace{-0.2cm}
\centering
\scalebox{0.7}{
\begin{tabular}{c|ccc}
\toprule
\multicolumn{2}{c|}{\textbf{Fully-supervised Method}} & \textbf{TVSum}   &    \textbf{SumMe}    \\ 
\midrule
\multicolumn{2}{c|}{SASUM \cite{wei2018video}}    & 53.9       & 40.6                                           \\ 
\midrule
% \multicolumn{2}{c|}{vsLSTM \cite{zhang2016video}}   & 54.2         & 37.6                                \\ 
% \midrule
\multicolumn{2}{c|}{dppLSTM \cite{zhang2016video}}  & 54.7        & 38.6                       \\ 
\midrule
% \multicolumn{2}{c|}{ActionRanking \cite{elfeki2019video}}   & 56.3       & 40.1                          \\ 
% \multicolumn{2}{|c|}{FPVSF [58]}  & ---  & \multicolumn{1}{c|}{23}  & 41.9  & \multicolumn{1}{c|}{29}&    \\ \hline
% \multicolumn{2}{|c|}{vsLSTM+Att \cite{casas2019video}}    & $*$      & 43.2                          \\ 
% \multicolumn{2}{c|}{H-RNN \cite{zhao2017hierarchical}}   & 57.7        & 41.1                              \\ 
% \multicolumn{2}{c|}{DR-DSN$_{sup}$ \cite{zhou2018deep}}  & 58.1          & 42.1                               \\ 
% \multicolumn{2}{|c|}{DSSE [79]}    & 57.0  & \multicolumn{1}{c|}{23}  & ---  & \multicolumn{1}{c|}{29}&    \\ \hline
% \multicolumn{2}{|c|}{dppLSTM+Att \cite{casas2019video}}   & $*$    &     & 43.8        &                 \\ \hline
% \multicolumn{2}{|c|}{WS-HRL [60]}  & 58.4  & \multicolumn{1}{c|}{23}     & 43.6  & \multicolumn{1}{c|}{29}& \\ \hline
% \multicolumn{2}{c|}{UnpairedVSN$_{psup}$ \cite{rochan2019video}}   & 56.1         & 48.0                    \\ 
% \midrule
\multicolumn{2}{c|}{ActionRanking \cite{elfeki2019video}}   & 56.3       & 40.1                          \\ 
\midrule
% \multicolumn{2}{c|}{SUM-FCN \cite{rochan2018video}}  & 56.8        & 47.5                                    \\ 
% \midrule
% \multicolumn{2}{c|}{SF-CVS \cite{huang2019novel}}    & 58.0         & 46.0                                     \\ 
\multicolumn{2}{c|}{H-RNN \cite{zhao2017hierarchical}}   & 57.7        & 41.1                              \\ 
\midrule
% \multicolumn{2}{|c|}{MAVS \cite{feng2018extractive}}  & 66.8    &     & 40.3        &                           \\ \hline
% \multicolumn{2}{c|}{SF-CVS \cite{huang2019novel}}    & 58.0         & 46.0                                     \\ 
% \midrule
\multicolumn{2}{c|}{CRSum \cite{yuan2019spatiotemporal}}   & 58.0       & 47.3                                       \\
\midrule
% \multicolumn{2}{c|}{DR-DSN$_{sup}$ \cite{zhou2018deep}}  & 58.1          & 42.1                               \\ 
% \midrule
% \multicolumn{2}{c|}{SASUM$_{fullysup}$ \cite{wei2018video}}   & 58.2         & 45.3                                \\ 
% \midrule
% \multicolumn{2}{c|}{SUM-DeepLab \cite{rochan2018video}}    & 58.4         & 48.8                                  \\ 
% \midrule
% \multicolumn{2}{c|}{CSNet$_{sup}$ \cite{jung2019discriminative}}   & 58.5          & 48.6                        \\ 
% \multicolumn{2}{|c|}{DQSN [76]}    & 58.6   & \multicolumn{1}{c|}{23}   & ---     & \multicolumn{1}{c|}{29}&  \\hline
% \midrule
% \multicolumn{2}{c|}{PCDL$_{sup}$ \cite{zhao2019property}}    & 59.2       & 43.7                             \\ 
% \midrule
% \multicolumn{2}{c|}{A-AVS \cite{ji2019video}}    & 59.4       & 43.9                                             \\  
% \midrule
% \multicolumn{2}{c|}{ACGAN$_{sup}$ \cite{he2019unsupervised}}  & 59.4       & 47.2                                \\ 
% \midrule
% \multicolumn{2}{c|}{HSA-RNN \cite{zhao2018hsa}}  & 59.8           & 44.1                                          \\ 
% \midrule
% \multicolumn{2}{c|}{H-MAN \cite{liu2019learning}}   & 60.4          & 51.8        \\ 
% \midrule
% \multicolumn{2}{c|}{SMLD \cite{chu2019spatiotemporal}}    & 61.0          & 47.6       \\ 
% \midrule
\multicolumn{2}{c|}{M-AVS \cite{ji2019video}}   & 61.0       & 44.4                                           \\ 
\midrule
\multicolumn{2}{c|}{VASNet \cite{fajtl2018summarizing}}  & 61.4       & 49.7                                    \\ 
\midrule
% \multicolumn{2}{|c|}{SMN \cite{wang2019stacked}}    & 64.5  &   & \textbf{58.3} &                                        \\ \hline
\multicolumn{2}{c|}{iPTNet \cite{jiang2022joint}}    & 63.4    & 54.5                       \\ 
\midrule
\multicolumn{2}{c|}{DASP \cite{ji2020deep}}  & 63.6        & 45.5                                    \\ 
\midrule
% \rowcolor{mygray} \multicolumn{2}{c|}{\textbf{Ours}}    & \textbf{67.5}        & \textbf{\textcolor{red}{55.4}}             \\
\rowcolor{mygray} \multicolumn{2}{c|}{\textbf{Causalainer}}    & \textbf{67.5}        & \textbf{52.4}             \\
\bottomrule
\end{tabular}}
\vspace{-0.4cm}
\label{table:table3}
\end{table}

\begin{table}[t!]
    \caption{Comparison with the multi-modal state-of-the-art. The proposed method outperforms the existing multi-modal approaches. `-' denotes unavailability from previous work. 
    }
\vspace{-0.2cm}
\centering
\scalebox{0.7}{
\begin{tabular}{c|ccc}
\toprule
\multicolumn{2}{c|}{\textbf{Multi-modal Method}} & \textbf{TVSum}   &    \textbf{QueryVS}    \\ 
\midrule
\multicolumn{2}{c|}{DSSE \cite{yuan2017video}}    & 57.0     & -    \\
\midrule
\multicolumn{2}{c|}{QueryVS \cite{huang2020query}}    & -      & 41.4  \\
\midrule
\multicolumn{2}{c|}{DQSN \cite{zhou2018video}}    & 58.6     & - \\
\midrule
\multicolumn{2}{c|}{GPT2MVS \cite{huang2021gpt2mvs}}    & -       & 54.8  \\
\midrule
\rowcolor{mygray} \multicolumn{2}{c|}{\textbf{Causalainer}}   & \textbf{68.2}     & \textbf{55.5} \\
\bottomrule
\end{tabular}}
\vspace{-0.4cm}
\label{table:table4}
\end{table}

\begin{table}[ht]
    \caption{Comparison with weakly-supervised state-of-the-art methods. The performance of the proposed approach is better than the existing weakly-supervised method.
    }
\vspace{-0.2cm}
\centering
\scalebox{0.7}{
\begin{tabular}{c|cc}
\toprule
\multicolumn{2}{c|}{\textbf{Weakly-supervised Method}} & \textbf{TVSum}      \\ 
\midrule
\multicolumn{2}{c|}{Random summary}  & 54.4         \\
\midrule
\multicolumn{2}{c|}{WS-HRL \cite{chen2019weakly}}  & 58.4    \\ 
\midrule
\rowcolor{mygray} \multicolumn{2}{c|}{\textbf{Causalainer}} & \textbf{66.9} \\ 
\bottomrule
\end{tabular}}
\vspace{-0.3cm}
\label{table:table5}
\end{table}

\noindent\textbf{3.2 Evaluation and Analysis}
\label{section:section4-2}

\noindent\textbf{Evaluation protocol.}
Following existing works \cite{jiang2022joint,huang2021gpt2mvs,huang2020query,gygli2014creating,song2015tvsum}, we evaluate the proposed method under the same setting. TVSum, QueryVS, and SumMe datasets are randomly divided into five splits, respectively. For each of them, $80$\% of the dataset is used for training, and the remaining for evaluation. 
% We run the models five times and report the averaged results \cite{jiang2022joint,huang2021gpt2mvs,huang2020query}. 
$F_{1}$-score \cite{hripcsak2005agreement,gygli2014creating,song2015tvsum,jiang2022joint} is adopted to measure the matching degree of the generated video summaries $\mathbb{S}_{i}$ and the ground-truth video summaries $\hat{\mathbb{S}}_{i}$ for video $i$. 
% It is defined based on precision and recall. The precision $P$ and recall $R$ based on the temporal overlap between $\hat{\mathbb{S}}_{i}$ and $\mathbb{S}_{i}$ are calculated as follows:
% \begin{align}
%     P = \frac{|\mathbb{S}_{i} \cap \hat{\mathbb{S}}_{i}|}{|\mathbb{S}_{i}|}, R = \frac{|\mathbb{S}_{i} \cap \hat{\mathbb{S}}_{i}|}{|\hat{\mathbb{S}}_{i}|}, F_{1}=\frac{2PR}{P+R}.
% \end{align}
% For videos with multiple human-annotated video summaries, the calculation of metrics in \cite{jiang2022joint,huang2021gpt2mvs} is followed.

% $F_{1}$-score is a commonly used evaluation metric in multi-modal video summarization based on the QueryVS and TVSum datasets \cite{hripcsak2005agreement,gygli2014creating,song2015tvsum,jiang2022joint}. Hence, $F_{1}$-score is also used to quantify the model performance in this work, referring to Table \ref{table:table3}, Table \ref{table:table4}, Table \ref{table:table5}, Table \ref{table:table2}, and \textbf{Supplementary}. 

\begin{figure}[t!]
\begin{center}
\includegraphics[width=0.9\linewidth]{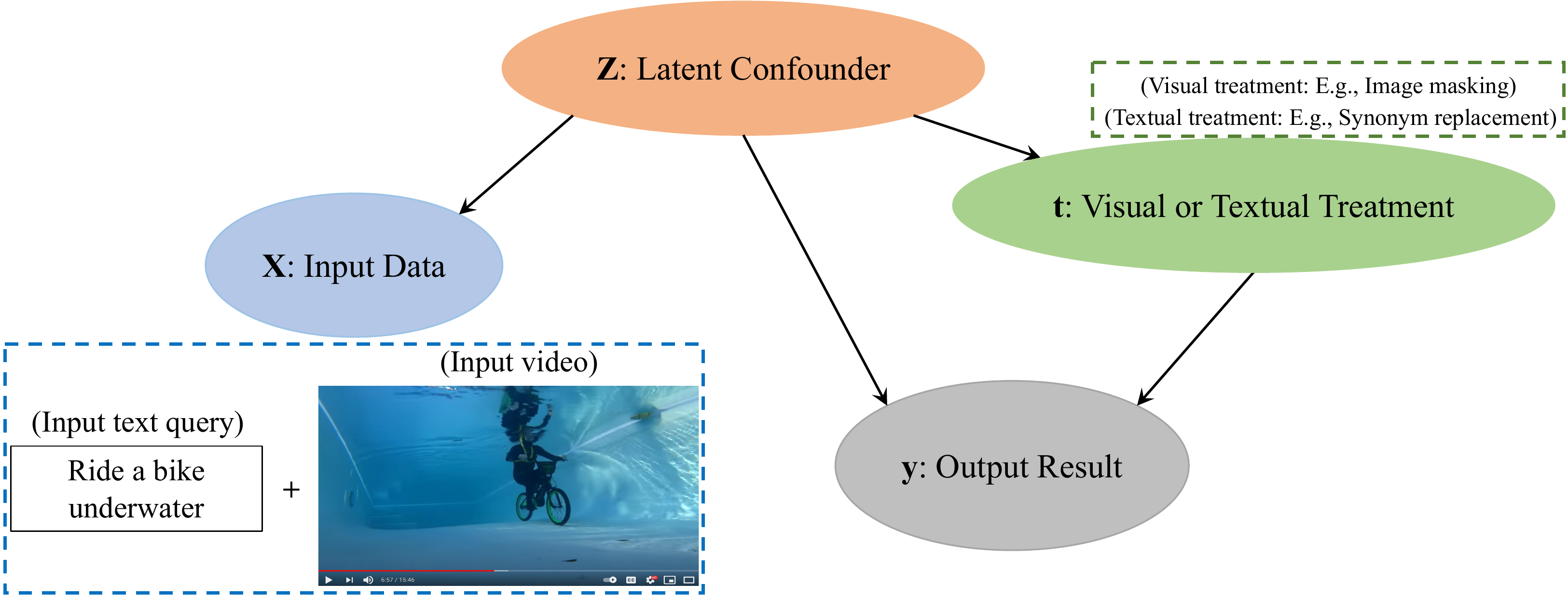}
\end{center}
\vspace{-0.60cm}
  \caption{Causal graph in video summarization. $\textbf{t}$ is an intervention, e.g., visual or textual perturbation. $\textbf{y}$ is an outcome, e.g., an importance score of a video frame or a relevance score between the input text query and video. $\textbf{Z}$ is an unobserved confounder, e.g., representativeness, interestingness, or storyline smoothness. $\textbf{X}$ is noisy views on the hidden confounder $\textbf{Z}$, say the input text query and video. The causality graph of video summarization leads to more explainable modeling.
  % \textcolor{red}{combine with Christ that 4 circle figure and redo this figure.}
  }
\vspace{-0.5cm}
\label{fig:figure39}
\end{figure}

\noindent\textbf{State-of-the-art comparisons.}
The proposed method outperforms existing state-of-the-art (SOTA) models based on different supervision schemes, as shown in Table \ref{table:table3}, Table \ref{table:table4}, and Table \ref{table:table5}. This is because the introduced causal modeling strengthens the causal inference ability of a video summarization model by uncovering the causal relations that guide the process and result.

\noindent\textbf{Effectiveness analysis of the proposed causal modeling.}
The proposed approach differs from existing methods by introducing causal modeling. Hence, the results in Tables \ref{table:table3}, \ref{table:table4}, and \ref{table:table5}, demonstrate the effectiveness of this approach and serve as an ablation study of causal learning. An auxiliary task/distribution is a key component of the proposed approach, helping the model learn to diagnose input to make correct inferences for the main task, i.e., video summary inference, During training, a binary causation label is provided to teach the model to perform well regardless of intervention. This implies the model has the ability to analyze input and perform well in the main task, making it more robust.

% Since the main difference between the proposed approach and the existing methods is the proposed causal modeling, the result in Table \ref{table:table3}, Table \ref{table:table4}, and Table \ref{table:table5} can be considered as an ablation study of the causal learning based on different supervision schemes. The results show that the introduced causal modeling is effective. One of the key components of the proposed causal learning is the auxiliary task/distribution. The purpose of the introduced auxiliary task is to help the model learn to diagnose the input to make the correct inference of the main task, i.e., video summary inference, despite the irrelevant intervention that exists. In the training phase, the ground truth binary causation label is provided to teach the model whether the input has interfered or not. If a model can do well no matter whether the intervention exists or not, it means the model has the ability to analyze the input in order to perform well in the main task. In other words, it implies the model has causal inference ability. In some sense, we also can say the proposed causal modeling makes a model become more robust.

\noindent\textbf{Explainability improvement analysis.}
The Causalainer method benefits modeling explainability with its associated causal graph of video summarization. Latent factors affecting video summary generation are treated as the causal effect in the proposed causal modeling. A causal graphical model is used to approach the video summarization problem, and the modeling explainability is illustrated in Figure \ref{fig:figure39}.

\section{Conclusion}
% Automated ML-based decision-making systems, e.g., video summarization, are not fully trusted because of lacking explainability. Since humans do not trust blindly, they want to understand a decision, or at least they want to get an explanation for certain decisions made by an ML-based model. In this work, a more explainable Causalainer method for video summarization is proposed. The associated causal graph for video summarization shows that the introduced causal modeling improves modeling explainability. The experimental results show that the proposed approach achieves state-of-the-art performance in terms of the $F_{1}$-score. 

% The lack of explainability in ML-based decision-making systems, such as video summarization, leads to mistrust. To address this issue, a more explainable Causalainer method is proposed in this work, which improves modeling explainability. The proposed method achieves state-of-the-art performance, in terms of the $F_{1}$-score, in video summarization, according to the experimental results.

ML-based decision-making systems, like video summarization, suffer from a lack of explainability, resulting in mistrust. To improve modeling explainability, we propose a new Causalainer method that achieves state-of-the-art $F_{1}$-score performance in video summarization.

{\small
\bibliographystyle{ieee_fullname}
\bibliography{ref}

\begin{thebibliography}{10}\itemsep=-1pt

\bibitem{abbasnejad2020counterfactual}
Ehsan Abbasnejad, Damien Teney, Amin Parvaneh, Javen Shi, and Anton van~den
  Hengel.
\newblock Counterfactual vision and language learning.
\newblock In {\em Proceedings of the IEEE/CVF Conference on Computer Vision and
  Pattern Recognition}, pages 10044--10054, 2020.

\bibitem{agarwal2020towards}
Vedika Agarwal, Rakshith Shetty, and Mario Fritz.
\newblock Towards causal vqa: Revealing and reducing spurious correlations by
  invariant and covariant semantic editing.
\newblock In {\em Proceedings of the IEEE/CVF Conference on Computer Vision and
  Pattern Recognition}, pages 9690--9698, 2020.

\bibitem{apostolidis2021video}
Evlampios Apostolidis, Eleni Adamantidou, Alexandros~I Metsai, Vasileios
  Mezaris, and Ioannis Patras.
\newblock Video summarization using deep neural networks: A survey.
\newblock {\em arXiv preprint arXiv:2101.06072}, 2021.

\bibitem{cai2018weakly}
Sijia Cai, Wangmeng Zuo, Larry~S Davis, and Lei Zhang.
\newblock Weakly-supervised video summarization using variational
  encoder-decoder and web prior.
\newblock In {\em Proceedings of the European Conference on Computer Vision
  (ECCV)}, pages 184--200, 2018.

\bibitem{cai2012identifying}
Zhihong Cai and Manabu Kuroki.
\newblock On identifying total effects in the presence of latent variables and
  selection bias.
\newblock {\em arXiv preprint arXiv:1206.3239}, 2012.

\bibitem{chen2019weakly}
Yiyan Chen, Li Tao, Xueting Wang, and Toshihiko Yamasaki.
\newblock Weakly supervised video summarization by hierarchical reinforcement
  learning.
\newblock In {\em Proceedings of the ACM Multimedia Asia}, 2019.

\bibitem{di2021dawn}
Riccardo Di~Sipio, Jia-Hong Huang, Samuel Yen-Chi Chen, Stefano Mangini, and
  Marcel Worring.
\newblock The dawn of quantum natural language processing.
\newblock {\em ICASSP}, 2022.

\bibitem{elfeki2019video}
Mohamed Elfeki and Ali Borji.
\newblock Video summarization via actionness ranking.
\newblock In {\em 2019 IEEE Winter Conference on Applications of Computer
  Vision (WACV)}, pages 754--763. IEEE, 2019.

\bibitem{fajtl2018summarizing}
Jiri Fajtl, Hajar~Sadeghi Sokeh, Vasileios Argyriou, Dorothy Monekosso, and
  Paolo Remagnino.
\newblock Summarizing videos with attention.
\newblock In {\em Asian Conference on Computer Vision}, pages 39--54. Springer,
  2018.

\bibitem{gong2014diverse}
Boqing Gong, Wei-Lun Chao, Kristen Grauman, and Fei Sha.
\newblock Diverse sequential subset selection for supervised video
  summarization.
\newblock In {\em Advances in neural information processing systems}, pages
  2069--2077, 2014.

\bibitem{greenland1983correcting}
Sander Greenland and DAVID~G KLEINBAUM.
\newblock Correcting for misclassification in two-way tables and matched-pair
  studies.
\newblock {\em International Journal of Epidemiology}, 12(1):93--97, 1983.

\bibitem{greenland2011bias}
Sander Greenland and Timothy~L Lash.
\newblock Bias analysis.
\newblock {\em International Encyclopedia of Statistical Science}, 2:145--148,
  2011.

\bibitem{gygli2014creating}
Michael Gygli, Helmut Grabner, Hayko Riemenschneider, and Luc Van~Gool.
\newblock Creating summaries from user videos.
\newblock In {\em ECCV}, pages 505--520. Springer, 2014.

\bibitem{gygli2015video}
Michael Gygli, Helmut Grabner, and Luc Van~Gool.
\newblock Video summarization by learning submodular mixtures of objectives.
\newblock In {\em Proceedings of the IEEE conference on computer vision and
  pattern recognition}, pages 3090--3098, 2015.

\bibitem{hara2018can}
Kensho Hara, Hirokatsu Kataoka, and Yutaka Satoh.
\newblock Can spatiotemporal 3d cnns retrace the history of 2d cnns and
  imagenet?
\newblock In {\em Proceedings of the IEEE conference on Computer Vision and
  Pattern Recognition}, pages 6546--6555, 2018.

\bibitem{he2016deep}
Kaiming He, Xiangyu Zhang, Shaoqing Ren, and Jian Sun.
\newblock Deep residual learning for image recognition.
\newblock In {\em Proceedings of the IEEE conference on computer vision and
  pattern recognition}, pages 770--778, 2016.

\bibitem{ho2018summarizing}
Hsuan-I Ho, Wei-Chen Chiu, and Yu-Chiang~Frank Wang.
\newblock Summarizing first-person videos from third persons' points of view.
\newblock In {\em Proceedings of the European Conference on Computer Vision
  (ECCV)}, pages 70--85, 2018.

\bibitem{hripcsak2005agreement}
George Hripcsak and Adam~S Rothschild.
\newblock Agreement, the f-measure, and reliability in information retrieval.
\newblock {\em Journal of the American medical informatics association},
  12(3):296--298, 2005.

\bibitem{hu2019silco}
Tao Hu, Pascal Mettes, Jia-Hong Huang, and Cees~GM Snoek.
\newblock Silco: Show a few images, localize the common object.
\newblock In {\em ICCV}, pages 5067--5076, 2019.

\bibitem{huang2017robustnessMS}
Jia-Hong Huang.
\newblock Robustness analysis of visual question answering models by basic
  questions.
\newblock {\em King Abdullah University of Science and Technology, Master
  Thesis}, 2017.

\bibitem{huang2017vqabq}
Jia-Hong Huang, Modar Alfadly, and Bernard Ghanem.
\newblock Vqabq: Visual question answering by basic questions.
\newblock {\em VQA Challenge Workshop, CVPR}, 2017.

\bibitem{huang2017robustness}
Jia-Hong Huang, Modar Alfadly, and Bernard Ghanem.
\newblock Robustness analysis of visual qa models by basic questions.
\newblock {\em VQA Challenge and Visual Dialog Workshop, CVPR}, 2018.

\bibitem{huang2019assessing}
Jia-Hong Huang, Modar Alfadly, Bernard Ghanem, and Marcel Worring.
\newblock Assessing the robustness of visual question answering.
\newblock {\em arXiv preprint arXiv:1912.01452}, 2019.

\bibitem{huang2023improving}
Jia-Hong Huang, Modar Alfadly, Bernard Ghanem, and Marcel Worring.
\newblock Improving visual question answering models through robustness
  analysis and in-context learning with a chain of basic questions.
\newblock {\em arXiv preprint arXiv:2304.03147}, 2023.

\bibitem{huang2019novel}
Jia-Hong Huang, Cuong~Duc Dao, Modar Alfadly, and Bernard Ghanem.
\newblock A novel framework for robustness analysis of visual qa models.
\newblock In {\em Proceedings of the Thirty-Third AAAI Conference on Artificial
  Intelligence}, volume~33, pages 8449--8456, 2019.

\bibitem{huang2021gpt2mvs}
Jia-Hong Huang, Luka Murn, Marta Mrak, and Marcel Worring.
\newblock Gpt2mvs: Generative pre-trained transformer-2 for multi-modal video
  summarization.
\newblock In {\em ICMR}, pages 580--589, 2021.

\bibitem{huang2020query}
Jia-Hong Huang and Marcel Worring.
\newblock Query-controllable video summarization.
\newblock In {\em ICMR}, pages 242--250, 2020.

\bibitem{huang2021contextualized}
Jia-Hong Huang, Ting-Wei Wu, and Marcel Worring.
\newblock Contextualized keyword representations for multi-modal retinal image
  captioning.
\newblock In {\em ICMR}, pages 645--652, 2021.

\bibitem{huang2022non}
Jia-Hong Huang, Ting-Wei Wu, C-H~Huck Yang, Zenglin Shi, I Lin, Jesper Tegner,
  Marcel Worring, et~al.
\newblock Non-local attention improves description generation for retinal
  images.
\newblock In {\em WACV}, pages 1606--1615, 2022.

\bibitem{huang2021deep}
Jia-Hong Huang, Ting-Wei Wu, Chao-Han~Huck Yang, and Marcel Worring.
\newblock Deep context-encoding network for retinal image captioning.
\newblock In {\em ICIP}, pages 3762--3766. IEEE, 2021.

\bibitem{huang2021longer}
Jia-Hong Huang, Ting-Wei Wu, Chao-Han~Huck Yang, and Marcel Worring.
\newblock Longer version for" deep context-encoding network for retinal image
  captioning".
\newblock {\em arXiv preprint arXiv:2105.14538}, 2021.

\bibitem{huang2022causal}
Jia-Hong Huang, Chao-Han~Huck Yang, Pin-Yu Chen, Andrew Brown, and Marcel
  Worring.
\newblock Causal video summarizer for video exploration.
\newblock In {\em 2022 IEEE International Conference on Multimedia and Expo
  (ICME)}, pages 1--6. IEEE, 2022.

\bibitem{huang2021deepopht}
Jia-Hong Huang, C-H~Huck Yang, Fangyu Liu, Meng Tian, Yi-Chieh Liu, Ting-Wei
  Wu, I Lin, Kang Wang, Hiromasa Morikawa, Hernghua Chang, et~al.
\newblock Deepopht: medical report generation for retinal images via deep
  models and visual explanation.
\newblock In {\em WACV}, pages 2442--2452, 2021.

\bibitem{huck2018auto}
C-H Huck~Yang, Fangyu Liu, Jia-Hong Huang, Meng Tian, I-Hung Lin, Yi~Chieh Liu,
  Hiromasa Morikawa, Hao-Hsiang Yang, and Jesper Tegner.
\newblock Auto-classification of retinal diseases in the limit of sparse data
  using a two-streams machine learning model.
\newblock In {\em ACCV}, pages 323--338. Springer, 2018.

\bibitem{ji2020deep}
Zhong Ji, Fang Jiao, Yanwei Pang, and Ling Shao.
\newblock Deep attentive and semantic preserving video summarization.
\newblock {\em Neurocomputing}, 405:200--207, 2020.

\bibitem{ji2019video}
Zhong Ji, Kailin Xiong, Yanwei Pang, and Xuelong Li.
\newblock Video summarization with attention-based encoder-decoder networks.
\newblock {\em IEEE Transactions on Circuits and Systems for Video Technology},
  2019.

\bibitem{jiang2022joint}
Hao Jiang and Yadong Mu.
\newblock Joint video summarization and moment localization by cross-task
  sample transfer.
\newblock In {\em Proceedings of the IEEE/CVF Conference on Computer Vision and
  Pattern Recognition}, pages 16388--16398, 2022.

\bibitem{jiang2019comprehensive}
Yudong Jiang, Kaixu Cui, Bo Peng, and Changliang Xu.
\newblock Comprehensive video understanding: Video summarization with
  content-based video recommender design.
\newblock In {\em Proceedings of the IEEE/CVF International Conference on
  Computer Vision Workshops}, pages 0--0, 2019.

\bibitem{khemakhem2021causal}
Ilyes Khemakhem, Ricardo Monti, Robert Leech, and Aapo Hyvarinen.
\newblock Causal autoregressive flows.
\newblock In {\em International Conference on Artificial Intelligence and
  Statistics}, pages 3520--3528. PMLR, 2021.

\bibitem{kingma2014adam}
Diederik~P Kingma and Jimmy Ba.
\newblock Adam: A method for stochastic optimization.
\newblock {\em arXiv preprint arXiv:1412.6980}, 2014.

\bibitem{kingma2013auto}
Diederik~P Kingma and Max Welling.
\newblock Auto-encoding variational bayes.
\newblock {\em arXiv preprint arXiv:1312.6114}, 2013.

\bibitem{lei2018action}
Jie Lei, Qiao Luan, Xinhui Song, Xiao Liu, Dapeng Tao, and Mingli Song.
\newblock Action parsing-driven video summarization based on reinforcement
  learning.
\newblock {\em IEEE Transactions on Circuits and Systems for Video Technology},
  29(7):2126--2137, 2018.

\bibitem{li2017extracting}
Yujie Li, Atsunori Kanemura, Hideki Asoh, Taiki Miyanishi, and Motoaki
  Kawanabe.
\newblock Extracting key frames from first-person videos in the common space of
  multiple sensors.
\newblock In {\em 2017 IEEE International Conference on Image Processing
  (ICIP)}, pages 3993--3997. IEEE, 2017.

\bibitem{liu2018synthesizing}
Yi-Chieh Liu, Hao-Hsiang Yang, C-H Huck~Yang, Jia-Hong Huang, Meng Tian,
  Hiromasa Morikawa, Yi-Chang~James Tsai, and Jesper Tegner.
\newblock Synthesizing new retinal symptom images by multiple generative
  models.
\newblock In {\em ACCV}, pages 235--250. Springer, 2018.

\bibitem{louizos2017causal}
Christos Louizos, Uri Shalit, Joris~M Mooij, David Sontag, Richard Zemel, and
  Max Welling.
\newblock Causal effect inference with deep latent-variable models.
\newblock In {\em NIPS}, pages 6446--6456, 2017.

\bibitem{otani2016video}
Mayu Otani, Yuta Nakashima, Esa Rahtu, Janne Heikkil{\"a}, and Naokazu Yokoya.
\newblock Video summarization using deep semantic features.
\newblock In {\em Asian Conference on Computer Vision}, pages 361--377.
  Springer, 2016.

\bibitem{panda2017weakly}
Rameswar Panda, Abir Das, Ziyan Wu, Jan Ernst, and Amit~K Roy-Chowdhury.
\newblock Weakly supervised summarization of web videos.
\newblock In {\em Proceedings of the IEEE International Conference on Computer
  Vision}, pages 3657--3666, 2017.

\bibitem{pearl2001bayesianism}
Judea Pearl.
\newblock Bayesianism and causality, or, why i am only a half-bayesian.
\newblock In {\em Foundations of bayesianism}, pages 19--36. Springer, 2001.

\bibitem{pearl2018theoretical}
Judea Pearl.
\newblock Theoretical impediments to machine learning with seven sparks from
  the causal revolution.
\newblock {\em arXiv preprint arXiv:1801.04016}, 2018.

\bibitem{rezende2014stochastic}
Danilo~Jimenez Rezende, Shakir Mohamed, and Daan Wierstra.
\newblock Stochastic backpropagation and approximate inference in deep
  generative models.
\newblock {\em arXiv preprint arXiv:1401.4082}, 2014.

\bibitem{sanabria2019deep}
Melissa Sanabria, Fr{\'e}d{\'e}ric Precioso, and Thomas Menguy.
\newblock A deep architecture for multimodal summarization of soccer games.
\newblock In {\em Proceedings Proceedings of the 2nd International Workshop on
  Multimedia Content Analysis in Sports}, pages 16--24, 2019.

\bibitem{selen1986adjusting}
Jan Sel{\'e}n.
\newblock Adjusting for errors in classification and measurement in the
  analysis of partly and purely categorical data.
\newblock {\em Journal of the American Statistical Association},
  81(393):75--81, 1986.

\bibitem{shalit2017estimating}
Uri Shalit, Fredrik~D Johansson, and David Sontag.
\newblock Estimating individual treatment effect: generalization bounds and
  algorithms.
\newblock In {\em International Conference on Machine Learning}, pages
  3076--3085. PMLR, 2017.

\bibitem{song2016category}
Xinhui Song, Ke Chen, Jie Lei, Li Sun, Zhiyuan Wang, Lei Xie, and Mingli Song.
\newblock Category driven deep recurrent neural network for video
  summarization.
\newblock In {\em 2016 IEEE International Conference on Multimedia \& Expo
  Workshops (ICMEW)}, pages 1--6. IEEE, 2016.

\bibitem{song2015tvsum}
Yale Song, Jordi Vallmitjana, Amanda Stent, and Alejandro Jaimes.
\newblock Tvsum: Summarizing web videos using titles.
\newblock In {\em Proceedings of the IEEE conference on computer vision and
  pattern recognition}, pages 5179--5187, 2015.

\bibitem{vasudevan2017query}
Arun~Balajee Vasudevan, Michael Gygli, Anna Volokitin, and Luc Van~Gool.
\newblock Query-adaptive video summarization via quality-aware relevance
  estimation.
\newblock In {\em Proceedings of the 25th ACM international conference on
  Multimedia}, pages 582--590, 2017.

\bibitem{vaswani2017attention}
Ashish Vaswani, Noam Shazeer, Niki Parmar, Jakob Uszkoreit, Llion Jones,
  Aidan~N Gomez, Lukasz Kaiser, and Illia Polosukhin.
\newblock Attention is all you need.
\newblock {\em arXiv preprint arXiv:1706.03762}, 2017.

\bibitem{wei2018video}
Huawei Wei, Bingbing Ni, Yichao Yan, Huanyu Yu, Xiaokang Yang, and Chen Yao.
\newblock Video summarization via semantic attended networks.
\newblock In {\em Proceedings of the AAAI Conference on Artificial
  Intelligence}, volume~32, 2018.

\bibitem{wu2023expert}
Ting-Wei Wu, Jia-Hong Huang, Joseph Lin, and Marcel Worring.
\newblock Expert-defined keywords improve interpretability of retinal image
  captioning.
\newblock In {\em Proceedings of the IEEE/CVF Winter Conference on Applications
  of Computer Vision}, pages 1859--1868, 2023.

\bibitem{yan2020self}
Xiang Yan, Syed~Zulqarnain Gilani, Mingtao Feng, Liang Zhang, Hanlin Qin, and
  Ajmal Mian.
\newblock Self-supervised learning to detect key frames in videos.
\newblock {\em Sensors}, 20(23):6941, 2020.

\bibitem{yang2018novel}
C-H~Huck Yang, Jia-Hong Huang, Fangyu Liu, Fang-Yi Chiu, Mengya Gao, Weifeng
  Lyu, Jesper Tegner, et~al.
\newblock A novel hybrid machine learning model for auto-classification of
  retinal diseases.
\newblock {\em Workshop on Computational Biology, ICML}, 2018.

\bibitem{yang2021causal}
Chao-Han~Huck Yang, I Hung, Te Danny, Yi Ouyang, and Pin-Yu Chen.
\newblock Causal inference q-network: Toward resilient reinforcement learning.
\newblock {\em arXiv preprint arXiv:2102.09677}, 2021.

\bibitem{yang2023treatment}
Chao-Han~Huck Yang, I-Te Hung, Yi-Chieh Liu, and Pin-Yu Chen.
\newblock Treatment learning causal transformer for noisy image classification.
\newblock In {\em Proceedings of the IEEE/CVF Winter Conference on Applications
  of Computer Vision}, pages 6139--6150, 2023.

\bibitem{yuan2019spatiotemporal}
Yuan Yuan, Haopeng Li, and Qi Wang.
\newblock Spatiotemporal modeling for video summarization using convolutional
  recurrent neural network.
\newblock {\em IEEE Access}, 7:64676--64685, 2019.

\bibitem{yuan2017video}
Yitian Yuan, Tao Mei, Peng Cui, and Wenwu Zhu.
\newblock Video summarization by learning deep side semantic embedding.
\newblock {\em IEEE Transactions on Circuits and Systems for Video Technology},
  29(1):226--237, 2017.

\bibitem{zhang2018advances}
Cheng Zhang, Judith B{\"u}tepage, Hedvig Kjellstr{\"o}m, and Stephan Mandt.
\newblock Advances in variational inference.
\newblock {\em IEEE transactions on pattern analysis and machine intelligence},
  41(8):2008--2026, 2018.

\bibitem{zhang2020causal}
Dong Zhang, Hanwang Zhang, Jinhui Tang, Xian-Sheng Hua, and Qianru Sun.
\newblock Causal intervention for weakly-supervised semantic segmentation.
\newblock {\em Advances in Neural Information Processing Systems}, 33, 2020.

\bibitem{zhang2016video}
Ke Zhang, Wei-Lun Chao, Fei Sha, and Kristen Grauman.
\newblock Video summarization with long short-term memory.
\newblock In {\em ECCV}, pages 766--782. Springer, 2016.

\bibitem{zhang2019dtr}
Yujia Zhang, Michael Kampffmeyer, Xiaoguang Zhao, and Min Tan.
\newblock Dtr-gan: Dilated temporal relational adversarial network for video
  summarization.
\newblock In {\em Proceedings of the ACM Turing Celebration Conference-China},
  pages 1--6, 2019.

\bibitem{zhao2017hierarchical}
Bin Zhao, Xuelong Li, and Xiaoqiang Lu.
\newblock Hierarchical recurrent neural network for video summarization.
\newblock In {\em Proceedings of the 25th ACM international conference on
  Multimedia}, pages 863--871, 2017.

\bibitem{zhao2018hsa}
Bin Zhao, Xuelong Li, and Xiaoqiang Lu.
\newblock Hsa-rnn: Hierarchical structure-adaptive rnn for video summarization.
\newblock In {\em Proceedings of the IEEE Conference on Computer Vision and
  Pattern Recognition}, pages 7405--7414, 2018.

\bibitem{zhou2018video}
Kaiyang Zhou, Tao Xiang, and Andrea Cavallaro.
\newblock Video summarisation by classification with deep reinforcement
  learning.
\newblock {\em arXiv preprint arXiv:1807.03089}, 2018.

\end{thebibliography}
}

\end{document}